\documentclass[letterpaper, 10 pt, conference]{ieeeconf}

\IEEEoverridecommandlockouts

\usepackage{microtype}
\usepackage{amsmath}
\usepackage{amssymb}
\usepackage{graphics}
\usepackage[tight]{subfigure}

\usepackage{tabularx}
\usepackage{booktabs}
\usepackage{multirow}
\usepackage{xspace}
\usepackage[ancient]{flushend}
\usepackage{algorithm}
\usepackage{algorithmic}

\usepackage{color}
\usepackage[textsize=scriptsize]{todonotes}
\usepackage{hyperref}
\usepackage{url}

\makeatletter
\let\NAT@parse\undefined
\makeatother

\usepackage[numbers,sort&compress]{natbib}
\bibpunct{[}{]}{,}{n}{}{;}

\title{Jointly Learning to Construct and Control Agents\\ using Deep Reinforcement Learning}

\author {Charles Schaff \hspace{20pt} David Yunis \hspace{20pt} Ayan Chakrabarti \hspace{20pt} Matthew R.\ Walter
\thanks{Charles Schaff and Matthew R.\ Walter are with the Toyota Technological Institute at Chicago, Chicago, IL USA, {\tt\small \{cbschaff,mwalter\}@ttic.edu}}
\thanks{David Yunis is with the Department of Mathematics at the University of Chicago, Chicago, IL USA, {\tt\small dyunis@uchicago.edu}}
\thanks{Ayan Chakrabarti is with the Department of Computer Science and Engineering at Washington University in St.\ Louis, St.\ Louis, MO USA, {\tt\small ayan@wustl.edu}}
}

\begin{document}

\maketitle

\begin{abstract}
  The physical design of a robot and the policy that controls its motion are inherently coupled, and should be determined according to the task and environment. In an increasing number of applications, data-driven and learning-based approaches, such as deep reinforcement learning, have proven effective at designing control policies. For most tasks, the only way to evaluate a physical design with respect to such control policies is empirical---i.e., by picking a design and training a control policy for it. Since training these policies is time-consuming, it is computationally infeasible to train separate policies for all possible designs as a means to identify the best one. In this work, we address this limitation by introducing a method that performs simultaneous joint optimization of the physical design and control network. Our approach maintains a distribution over designs and uses reinforcement learning to optimize a control policy to maximize expected reward over the design distribution. We give the controller access to design parameters to allow it to tailor its policy to each design in the distribution. Throughout training, we shift the distribution towards higher-performing designs, eventually converging to a design and control policy that are jointly optimal. We evaluate our approach in the context of legged locomotion, and demonstrate that it discovers novel designs and walking gaits, outperforming baselines in both performance and efficiency.
\end{abstract}

\section{Introduction} \label{sec:introduction}

A robot's ability to successfully interact with its environment depends both on its physical design as well as its proficiency at control, which are inherently coupled. Therefore, designing a robot requires reasoning both over the mechanical elements that make up its physical structure as well the control algorithm that regulates its motion.
These physical and computational design parameters of a robot must be optimized \emph{jointly}---different physical designs enable different control strategies, and the design process involves determining the optimal combination that is best suited to the robot's target task and environment.

Consider the development of a legged robot.
Different designs will have different optimal gaits, even when the morphology is preserved. Some designs may render locomotion impossible for a target environment (e.g., a robot with short legs  may be unable to locomote quickly), while others may make the underlying control problem easy to solve and naturally efficient (e.g., certain bipedal designs enable passive walking~\citep{mcgeer90,goswami98,collins01}). Rather than optimizing a robot's design or gait in isolation, it is therefore beneficial to consider them together as part of a joint optimization problem.
Thus, many researchers have explored approaches that jointly reason over physical design and control~\citep{digumarti14,ha17,spielberg17}. Most recent methods are aimed at ``model-based'' approaches to control---in that they require a model of the robot dynamics or a near-ideal motion trajectory, which is chosen based on expert intuition about a specific domain and task.

Data-driven and learning-based methods, such as deep reinforcement learning, have proven effective at designing control policies for an increasing number of tasks~\citep{tobin17,rajeswaran17,levine18}. However, unlike the above methods, most learning-based approaches do not admit straightforward analyses of the effect of changes to the physical design on the training process or the performance of their policies---indeed, learning-based methods may arrive at entirely different control strategies for different designs. Thus, the only way to evaluate the quality of different physical designs is by training a controller for each---essentially treating the physical design as a ``hyper-parameter'' for optimization. However, training controllers for most applications of even reasonable complexity is time-consuming. This makes it computationally infeasible to evaluate a diverse set of designs in order to determine which is optimal, even with sophisticated methods like Bayesian optimization to inform the set of designs to explore.

\begin{figure*}[!t]
    \centering
    \includegraphics[width=0.845\linewidth]{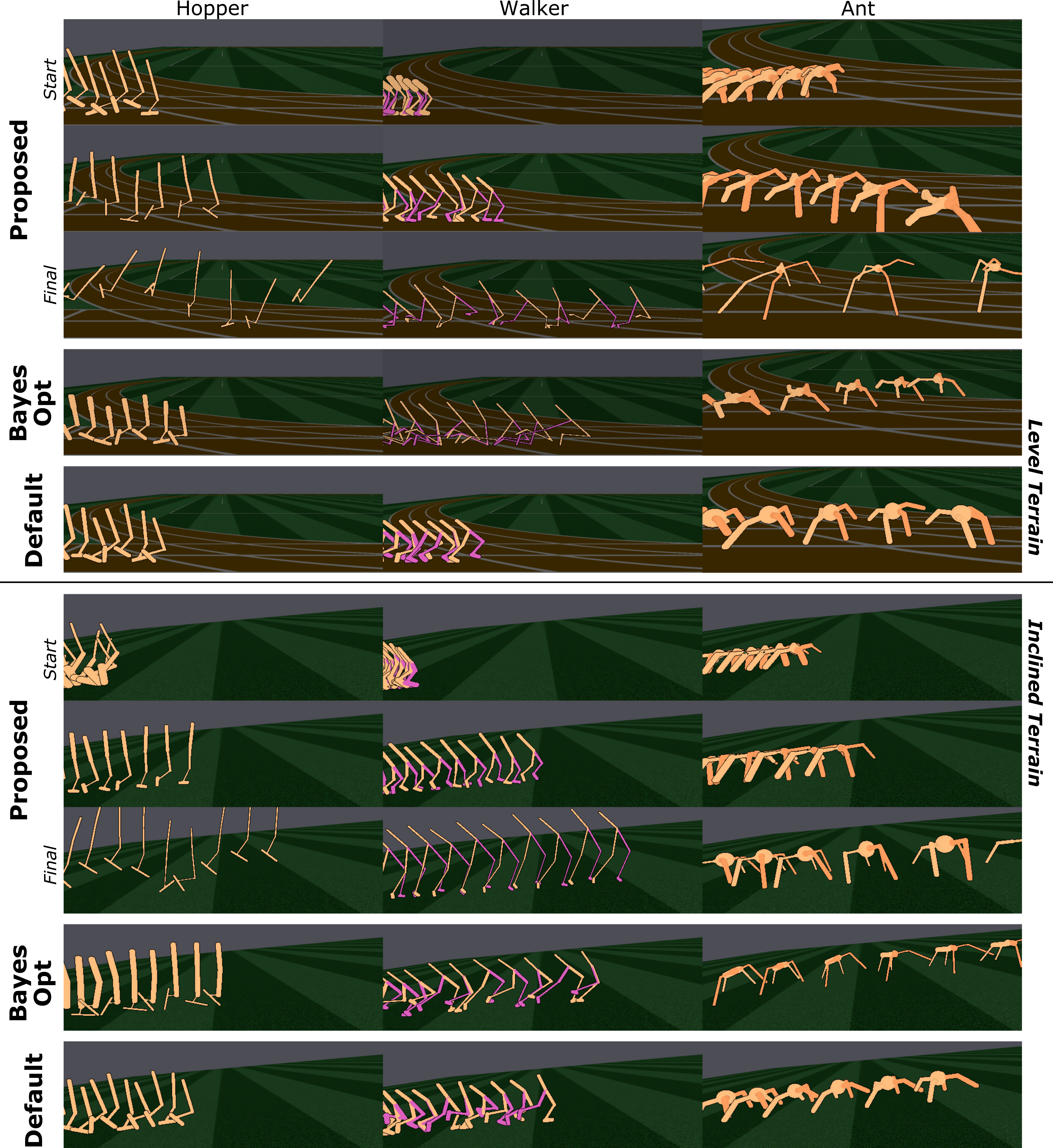}
    \caption{Our algorithm learns a robot's physical design jointly with the control policy to optimize performance for a given task and environment. Our method maintains a distribution over designs and optimizes a neural network control policy to maximize expected reward over this design distribution using reinforcement learning. The control network has access to the design parameters, allowing it to tailor its policy to each design in the distribution. We shift this distribution towards higher performing designs throughout training, eventually converging to a design and control policy that are jointly optimal. Here, we visualize the evolution of modes of the design distribution and the corresponding learned control policy (gait) for three robot morphologies tasked with locomoting over level (top) and inclined (bottom) terrains. We also show the default Roboschool designs with their learned gaits along with those learned using Bayesian optimization. Images are captured at fixed time intervals, so designs appearing farther to the right represent faster motion.}\label{fig:evolution-gaits}
\end{figure*}

In this work, we seek to alleviate these limitations by introducing an efficient algorithm that jointly optimizes over both physical design and control. Our approach maintains a distribution over designs and uses reinforcement learning to optimize a neural network control policy to maximize expected reward over the design distribution. We give the controller access to design parameters to allow it to tailor its policy to each design in the distribution. Throughout training, we shift the distribution towards higher performing designs and continue to train the controller to track the design distribution. In this way, our approach converges to a design and control policy that are jointly optimal. Figure~\ref{fig:evolution-gaits} visualizes this evolution for different robot morphologies.

Our approach can be applied to arbitrary morphologies, tasks, and environments, and explores the joint space of design and control in a purely data-driven fashion. We evaluate our method in the context of legged locomotion, parameterizing the length and radius of links for several different robot morphologies. Experimental results show that starting from random initializations, our approach consistently finds novel designs and gaits that exceed the performance of manually designed agents and other optimization baselines, across different morphologies and environments.

\section{Related Work} \label{sec:related-work}

Co-design of physical structure and control has a long history in robotics research. A large number existing approaches~\citep{park94,pil96,reyer02,villarreal13} are model-based---in that they rely on having a model of environment dynamics to solve for both the physical design and the control policy. Many focus on co-design with specific types of controllers, e.g., \citet{ha17} model both design and motion parameters via a set of implicit functions that express robot dynamics and actuation limits with a desired trajectory in mind, and carry out optimization on linearized approximations of these functions. Others are designed for settings in which control is formulated as a trajectory optimization problem~\cite{mordatch12, tassa12, wampler13, levine14, mordatch14, mordatch15, dai16, posa16, griffin16}. For example, \citet{spielberg17} solve for the design of articulated robots jointly with trajectory parameters (e.g., contact forces and torques), with the requirement that the problem be initialized within a small neighborhood of a feasible solution, and that the user provide an estimate of the robot configuration at each time step. Also worth noting are methods that find robot designs to meet specified task requirements---e.g., given user demonstrations of desired behaviors, \citet{coros13} learn optimal kinematic linkages capable of reproducing these motions, while \citet{mehta14} synthesize electro-mechanical designs in a compositional fashion based upon a complete, user-specified structural specification of the robot. Related, \citet{censi17} describes a theoretical framework that allows one to select discrete robot parts according to functional constraints, but does not reason over geometry or motion.

For many applications and domains, control policies modeled as neural networks and trained with deep reinforcement learning have emerged as successful approaches that deliver state-of-the-art performance~\citep{riedmiller05,mnih13,schulman15}. These techniques have been applied to control simple, simulated robots~\citep{wawrzynski09,watter15,lillicrap15}, robot manipulators~\citep{levine16}, and legged robots~\citep{schulman15a,peng16,tan18}. The ability of such controllers to successfully learn complex policies directly from low-level sensory input and without any expert supervision is clearly attractive. However, being purely data-driven and trained through a complex iterative process, the resulting control policies make it non-trivial to jointly optimize these policies with physical structure.

Early approaches to the co-design of neural controllers with physical designs~\cite{sims1994,lipson00,paul01,bongard11} employ evolutionary methods, albeit to restricted settings (e.g., \cite{bongard11} only optimizes over quadrupedal foothold patterns). Meanwhile, when dealing with deep neural networks that require a significant amount of computational expense to train, Bayesian optimization~\cite{bayesopt} has emerged as a successful strategy for the optimization of external (hyper-)parameters. This provides a possible strategy for learning physical designs by treating design specifications as hyper-parameters. However, Bayesian optimization also becomes computationally expensive and infeasible when applied to high-dimensional search spaces arising out of a large number of hyper-parameters.

Our work is motivated by the recent success of joint training of neural network-based estimators with sensor parameters: \citet{chakrabarti16} considers the problem of jointly learning a camera sensor's multiplexing pattern along with a neural network for reconstruction, while \citet{schaff17} optimize the placement of beacons in an environment jointly with a neural network for location estimation. However, since their loss and measurement functions are differentiable, these methods are able to rely on gradient-based updates, both for training the estimators and for computing gradients with respect to the sensor parameters. In this work, we also propose a joint optimization framework that trains both the physical design and a control policy network in a reinforcement learning setting based on rewards returned by the environment.

\newcommand{\mS}{\mathcal{S}}
\newcommand{\mA}{\mathcal{A}}
\newcommand{\mP}{\mathcal{P}}
\newcommand{\mR}{\mathcal{R}}
\newcommand{\mE}{\mathcal{E}}
\newcommand{\mM}{\mathcal{M}}

\section{Approach} \label{sec:approach}

We first review the standard reinforcement learning approach to learning a control policy for a given physical design. We then introduce our approach to simultaneously optimize the physical design and the policy as the latter is being trained.

\subsection{Control Policy Training  by Reinforcement Learning}

The problem of controlling an agent can be modeled as a continuous Markov decision process (MDP), denoted by the tuple $\{\mS,\mA,p,r, p_0\}$, where $\mS \subseteq \mathbb{R}^d$ is the state space, $\mA \subseteq \mathbb{R}^n$ the action space, $p$ the (unknown) transition model between states, $r$ the reward function, and $p_0$ the initial state distribution. An agent in state $s_t \in \mS$ at time $t$ takes action $a_t \in \mA$ and the environment returns the agent's new state $s_{t+1}$ according to the unknown transition function $p(s_{t+1} \vert s_t,a_t)$, along with the associated reward $r_t = r(s_t, a_t)$. The goal is to learn the control policy $\pi^*: \mS \rightarrow \mA$ mapping states to actions that
maximizes the expected reward $\mathbb{E}_\pi[\mR_t]$, which takes into account rewards at future time-steps  $\mR_t = \sum_{i=0}^\infty \gamma^i r_{t+i}$ with a discount factor $\gamma \in [0,1)$.

For an increasing number of reinforcement learning problems, a stochastic policy $\pi_\theta(a_t \vert s_t)$ is modeled as a neural network with input $s_t$ and output $a_t$. In the case of complex continuous action spaces, policy gradient methods are commonly used to learn the parameters $\theta$ through stochastic gradient ascent on the expected return.
``Vanilla'' policy gradient methods compute an estimate of the gradient $\nabla_\theta \mathbb{E}[\mR_t]$ using a sample-based mean computed over $\nabla_\theta \log \pi_\theta(a_t \vert s_t)\mR_t$~\citep{williams92}, which yields an unbiased gradient estimate~\citep{sutton00}.
Recently, methods have been proposed to improve the stability of the learning process~\cite{schulman15,schulman17}. Notably, proximal policy optimization (PPO)~\cite{schulman17} is a first-order class of methods that alternates between sampling data from the environment and optimizing the objective
\begin{equation}
    \hat{\mathbb{E}}_t\left[\min(r_t(\theta)\hat{A}_t, \text{clip}(r_t(\theta), 1-\epsilon, 1+\epsilon)\hat{A}_t)\right],
\end{equation}
where $r_t(\theta) = \frac{\pi_\theta(a_t \vert s_t)}{\pi_{\theta_{\text{old}}}(a_t \vert s_t)}$ and $\hat{\mathbb{E}}_t$ represents the average over a limited sample set. This clipped objective has the effect of maximizing expected return by making only small steps in policy space at a time.
This yields a simple yet robust reinforcement learning algorithm that attains state-of-the-art results on a wide array of tasks~\citep{schulman17}.

\subsection{Simultaneous Optimization of Physical Design and Control}

Standard reinforcement learning trains a policy $\pi_\theta$ for an agent with a fixed physical design. This design modulates the agent's interactions with the environment, specifically through the transition dynamics $p(s_{t+1} \vert s_t, a_t)$, and different designs will naturally have different optimal control policies $\pi^*_\theta$. The goal of our work is to enable the discovery of an optimal design $\omega^* \in \Omega$ from the space of feasible designs $\Omega$ that maximizes the agent's success when used in concert with a corresponding optimal control policy $\pi^*_\theta$.

Thus, we are interested in finding a design $\omega^*$ and corresponding policy $\pi^*_\theta(a_t \vert s_t, \omega^*)$
that are jointly optimal. Unfortunately, this search cannot be decoupled. Evaluating the quality of a design $\omega$ requires first determining the optimal policy $\pi^*_\theta$ for that design, which involves running a full round of training. In this respect, the design $\omega$ can be thought of as a ``hyper-parameter'', and the standard way of finding $\omega$ involves training different policies for different candidate designs. Unfortunately, this is computationally expensive particularly when the design space  $\Omega$ is high-dimensional, making it infeasible to sufficiently explore the space of designs, even with the aid of hyper-parameter optimization methods such as Bayesian optimization~\cite{bayesopt}.

In this work, instead of treating policy training as black-box optimization that we perform independently for each possible design, we propose searching over designs simultaneously with a search over policies. We extend the standard reinforcement learning formulation of control to also include $\omega$ as a learnable parameter. We let $\mE_\omega$ denote the MDP for a specific design $\omega$, such that all $\mE_\omega, \omega\in\Omega $ share a common state space $\mS$, action space $\mA$, reward function $r(s_t,a_t)$, and initial state distribution $p_0$. The MDPs differ only in their transition dynamics $p(s_{t+1} \vert s_t, a_t, \omega)$ according to their design $\omega$. Note that the requirement of a common action space limits $\Omega$ to robot designs that share a common morphology (same number of limbs, joints, etc.). Our goal is to find the optimal design $\omega^*$ and policy $\pi_{\theta}^*$ pair that maximizes expected reward.

We propose a framework for finding this optimal pair that, in addition to the policy distribution $\pi_\theta$, maintains a distribution $p_\phi(\omega)$ over possible designs $\omega\in\Omega$. The learnable parameters $\phi$ encode the framework's belief of which designs are likely to be successful. The policy function $\pi_\theta(a_t \vert s_t, \omega)$ is now provided with the parameters $\omega$ of the design it is controlling and trained to be successful with not just a single design, but multiple designs sampled from $p_\phi(\omega)$. Moreover, since we provide $\pi_\theta$ with design parameters, it is not constrained to use the same policy for all designs and is able to tailor control to each design.
\begin{figure}[!t]
    \centering
    \includegraphics[width=1.0\linewidth]{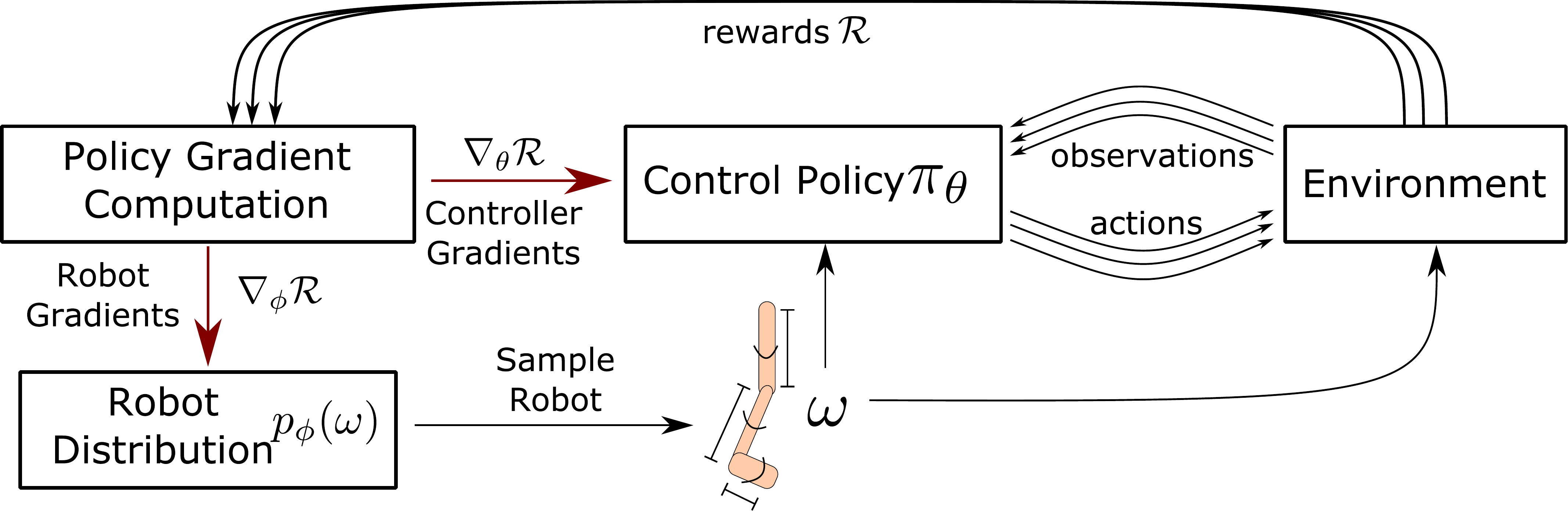}
    \caption{Our proposed algorithm maintains a distribution $p_\phi(\omega)$ over possible physical design parameters. At each iteration of training, a specific design $\omega$ is sampled from this distribution, and an episode is run with that design and the current control policy $\pi_{\theta}$ in the environment. The policy $\pi_\theta$ has access to the parameters $\omega$ of the robot instance it is controlling in that episode, along with observations fed back from the environment. Policy gradients with respect to episode rewards are then used to update parameters $\theta$ and $\phi$ of both the control policy and the robot distribution, respectively.} \label{fig:algorithm}
\end{figure}

Formally, we seek to find design and policy parameters $\phi^*$ and $\theta^*$ such that:
\begin{equation}
    \label{objective}
    \phi^*, \theta^*  = \underset{\phi, \theta}{\text{arg max }} \mathbb{E}_{\omega \sim p_\phi}\bigl[\mathbb{E}_{\pi_\theta}\left[\mR_t\right]\bigr].
\end{equation}
We propose an algorithm  (summarized in Algorithm~\ref{algorithm}) that carries out this optimization to maximize the expected reward obtained by the policy $\pi_\theta$ over the design distribution $p_\phi$.
\begin{algorithm}[!t]
  \caption{Joint Optimization of Design and Control} \label{algorithm}
  \begin{algorithmic}
      \STATE Initialize $\pi_\theta(a \vert s,\omega)$, $p_\phi(\omega)$, $T=0$
      \WHILE{True}
        \STATE Sample designs $\{\omega_1$,\ldots, $\omega_n\}$ s.t.
        $\omega_i \sim p_\phi$
        \STATE Control each design with $\pi_\theta$ for $t$ timesteps, collecting trajectories $\{s_1, a_1, r_1, \ldots, s_{t-1}, a_{t-1}, r_{t-1}, s_t\}_1, \ldots,$ $\{s_1, a_1, r_1, \ldots, s_{t-1}, a_{t-1}, r_{t-1}, s_t\}_n$
        \STATE Update $\theta$ using PPO.
        \STATE $T = T + nt$
        \IF{$T > C$}
            \STATE Compute average episode returns $R_1, \ldots, R_n$
            \STATE Update $\phi$ using $\nabla \phi \approx \frac{1}{n}\sum_{i=1}^n \nabla \text{log} p_\phi(\omega_i)R_i$
            \IF{$T$ mod $N == 0$ and \#components $> 1$}
                \STATE Sample and evaluate $100$ designs from each component of $p_\phi$.
                \STATE Remove the half of the components with the lowest average reward.
            \ENDIF
        \ENDIF
      \ENDWHILE
  \end{algorithmic}
\end{algorithm}

At each iteration of training, the policy is trained (using PPO) with respect to gradients that maximize the expected return over designs sampled from the current design distribution $p_\phi$. At the same time, the design distribution is updated every iteration to increase the probability density around designs that perform well when using the current learned policy $\pi_\theta$:
\begin{equation}
    \label{design_update}
    \nabla \mathbb{E}_{\omega \sim p_\phi}\bigl[\mathbb{E}_{\pi_\theta}\left[\mR_t\right]\bigr] = \mathbb{E}_{\omega \sim p_\phi}\bigl[\nabla \log p_\phi(\omega)\mathbb{E}_{\pi_\theta}\left[\mR_t\right]\bigr]
\end{equation}
This shifts the parameters of the design distribution $\phi$ to maximize the expected reward under the current policy $\pi_\theta$, and is analogous to gradient-based updates to the policy parameters~\cite{sehnke10}, except that the choice of $\phi$ now affects the transition dynamics. We use a Gaussian mixture model to parameterize the design distribution $p_\phi$, which allows our framework to maintain multiple distinct hypotheses for designs it believes to be promising. This allows greater exploration of the physical design space $\Omega$ and helps prevent the optimization from collapsing to local optima. Note that while the per-component means and variances are learned using gradient-based updates as described above, our framework uses a different approach to updating the mixing probabilities. It maintains a uniform distribution across components, and then eliminates the components whose samples yield the lowest reward at discrete intervals---halving the number of components every $N$ iterations.

At the beginning of the optimization process, the design distribution is initialized to have high entropy---with each component initialized with random means and high variance. Consequently, the policy initially learns to control a diverse set of designs. There is a warm-up period for the policy network at the beginning of training, where only the policy parameters $\theta$ are updated while the design parameters $\phi$ are kept fixed. Once the method begins to update the design distribution, the gradient updates proceed to eliminate designs that perform poorly as training continues.  This in turns allows the policy network to specialize, using its capacity to focus on a more restricted design set. Thus, the variance in the design distribution $p_\phi$ decreases and the expected reward increases as training progresses.  At the end of training, the design $\omega$ is fixed to the mode of the final estimate of the design distribution $p_\phi(\omega)$, and the policy network is fine-tuned for a few more iterations with this fixed design. This procedure yields the final estimates for the design $\omega^*$ and policy $\pi^*_\theta$.

\section{Experiments} \label{sec:results}

We evaluate our framework on the problem of jointly learning the physical design and control policy of legged robots tasked with locomoting within different environments. We consider three commonly used robot morphologies included in OpenAI's Roboschool~\citep{schulman17}: the Hopper, Walker (referred to as Walker2D in Roboschool), and Ant. We optimize both with the goal of maximizing a standard reward function, and find that our approach discovers novel robot designs and gaits that outperform controllers learned for the standard Roboschool design, as well as design-controller pairs learned through random sampling and Bayesian optimization. Source code and a video highlighting our results can be found at \url{http://ttic.uchicago.edu/~cbschaff/nlimb}.

\subsection{Experiment Setup}

We consider the task of locomotion for the Hopper, Walker, and Ant morphologies on both a level and inclined ground plane (with a five degree slope). The environments are built on top of Bullet Physics, a popular open-source physics engine. We use the Roboschool default reward function, which is a weighted sum of rewards for forward progress and staying upright, and penalties for torques and for reaching joint limits. Every episode ends when a robot falls over or after a maximum number of timesteps.

We restrict our method and the baselines to optimize over robot designs that adhere to the default Roboschool morphologies. We parameterize each morphology in terms of the length and radius (e.g., mass) of each link, and the spherical body radius for the ant. We impose symmetry over both legs of the Walker, but optimize over independent design parameters for each of the four legs of the Ant. Consequently, the Hopper and Walker both have eight learnable parameters, while the Ant has twenty-five. We limit the values of each parameter to lie within a reasonable range (as a proxy for fabrication constraints). Note the resulting search space includes a wide variety of robots with different shapes and sizes, including many that are clearly impractical for locomotion. However, rather than providing a reduced search space guided by expert intuition, we let our framework discover and reject such designs automatically.

The MDP state $s_t$ is comprised of joint angles and velocities, center-of-mass velocity, height of the torso, and direction to the target. For all experiments, we model the control policy $\pi_{\theta}(a_t \vert s_t, \omega)$ as a feed-forward neural network consisting of three fully-connected layers with $128$ units and $\tanh$ activation functions. A final layer maps the output to the robot action space---a vector of joint torques. For our framework, we also append the design parameters $\omega$ to the state variables before passing them to the control network.

For our framework, we represent the distribution over robot designs $p_{\phi}(\omega)$ as a GMM with eight mixture components, each with a diagonal covariance matrix. We randomly initialize the means of each component and set the variances such that the distribution spans the parameter ranges. The mixing probabilities are set to be uniform---across all eight components at the start of training, and across the remaining components after low reward components are removed every $N$ timesteps. Thus, only the component means and variances are updated based on policy gradients during training.

\begin{figure*}[!th]
    \centering
    \includegraphics[width=\linewidth]{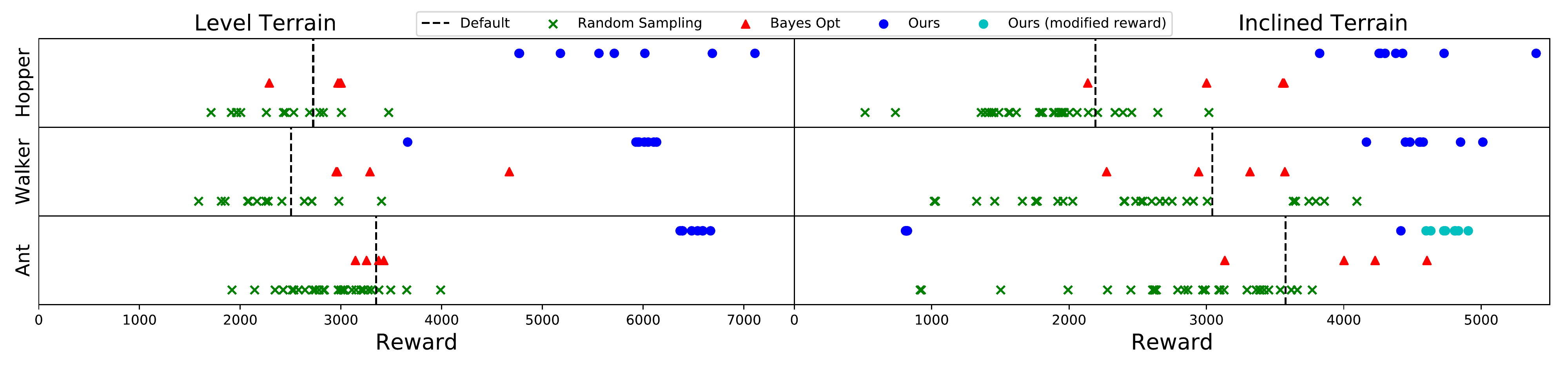}
    \caption{We show a comparison of the rewards achieved for different seeds with our method and the three baselines for level (left) and  inclined (right) terrains. Performance is measured as the average reward over $100$ episodes at the end of training. The cyan dots for the Ant on inclined terrain represent our performance after modifying the reward function for training, but their position indicates performance according to the original reward function.} \label{fig:rewards}
\end{figure*}

Our framework trains in parallel on eight robot samples and updates the policy $\pi_{\theta}(a_t \vert s_t, \omega)$ with PPO, for a total of $1$\,B environment timesteps. We train the controller for $100$\,M timesteps before updating the design distribution. In the case of the Ant, which takes longer to train, we train for a total of $1.5$\,B timesteps and start updating the design distribution after $200$\,M timesteps. Additionally, we evaluate and prune components every $N=100$\,M ($200$\,M for the Ant) timesteps. For the last $100$\,M timesteps, we choose the mean of $p_\phi$ as the optimal design $\omega^*$ and train the policy with this fixed design.

\subsection{Baselines}
In addition to reporting results from our framework (run eight times for each morphology and environment to gauge consistency), we evaluate and compare it to three baselines:

\subsubsection*{Default Designs} This baseline simply involves training a control policy for the standard, hand-crafted Roboschool designs. We use the same policy architecture as with our method and train until convergence.

\subsubsection*{Random Sampling} This baseline samples designs uniformly at random within the parameter ranges, and trains a separate control policy for each sample. We use the same policy architecture as with our method and train until convergence. Since this approach is inherently more parallelizable than our method, we allow this baseline to sample designs and train control policies for three times the number of timesteps used by our method. The reward of the best design-control pair found across samples can then be compared to the performance of a single run of our experiment.

\subsubsection*{Bayesian Optimization} This baseline employs Bayesian optimization (BayesOpt) to search jointly over the physical design space, using the implementation from \cite{gpyopt2016}. For each sampled design, we train a control policy until convergence and return the average episode reward to the BayesOpt routine. BayesOpt then samples a new design and the process repeats. We use the same policy architecture as our experiments and allow BayesOpt the same number of environment timesteps used by our method. However, this timestep limit only allows for roughly ten designs to be evaluated. To account for variance in outcomes, we run four copies of this baseline with different random seeds and report each outcome.

\subsection{Results}

Figure~\ref{fig:evolution-gaits} presents the physical designs and gaits learned by our method and BayesOpt, as well as the gait learned for the fixed design, for all three morphologies and both environments. For BayesOpt and our method, we show the best performing seed. Gaits are visualized by capturing images at fixed time intervals, so designs appearing farther to the right represent faster motion. For our method, in addition to showing the final learned design and gait, we also visualize results from near the beginning and middle of training to illustrate the evolution of the design and control policy. Figure~\ref{fig:rewards} provides a more quantitative analysis of the rewards achieved by all seeds of the different methods.

\subsubsection*{Level Terrain} We find that the joint design-policy pair found by our framework exceeds the performance of all baselines for all three morphologies on level terrain. Our method obtains these performance levels by discovering unique robot designs together with novel walking gaits. Note that our method learns these designs and control policies from a \emph{random initialization}, without access to a dynamics model, expert supervision, or good initial solutions.

For the Hopper, there is some variance in performance among the seeds for our method, but every seed outperforms all seeds of the different baselines. We find that our learned Hopper designs all have long bodies and generally use one of two gaits. Designs with a heavier body have short, quick strides, while designs with a narrow, light body have large, bounding gaits. For both groups, the longer torso of the learned robot improves stability, allowing it to maintain balance while locomoting at a faster pace.

The results for the Walker are more consistent, with all but one seed achieving similar performance. All of our seeds outperform the random sampling and standard design baselines. Compared to BayesOpt, our framework yields better results more consistently---seven of our seeds outperform all designs learned through BayesOpt with one seed doing worse than one of the BayesOpt seeds. The seven better-performing designs from our framework all have long thin legs and small feet, and all achieve a running gait with long strides. The last seed learns a design with long feet and even longer legs, and learns a stable walking gait on its knees. The low stance enables the Walker to fully extend its leg backwards, creating a long stride similar to that of a sprinter using starting blocks.

Our method achieves consistent results for the Ant morphology, all of which do better than the baselines by a significant margin. The learned design has a small, lightweight body and thin legs. Despite not sharing parameters between legs, our method consistently finds a solution in which each leg is roughly the same size and shape. The learned gait primarily uses three legs (one on each side and one behind) to walk and generally keeps the front leg off the ground, occasionally using it for balance.

\begin{figure*}[!th]
    \centering
    \includegraphics[width=\linewidth]{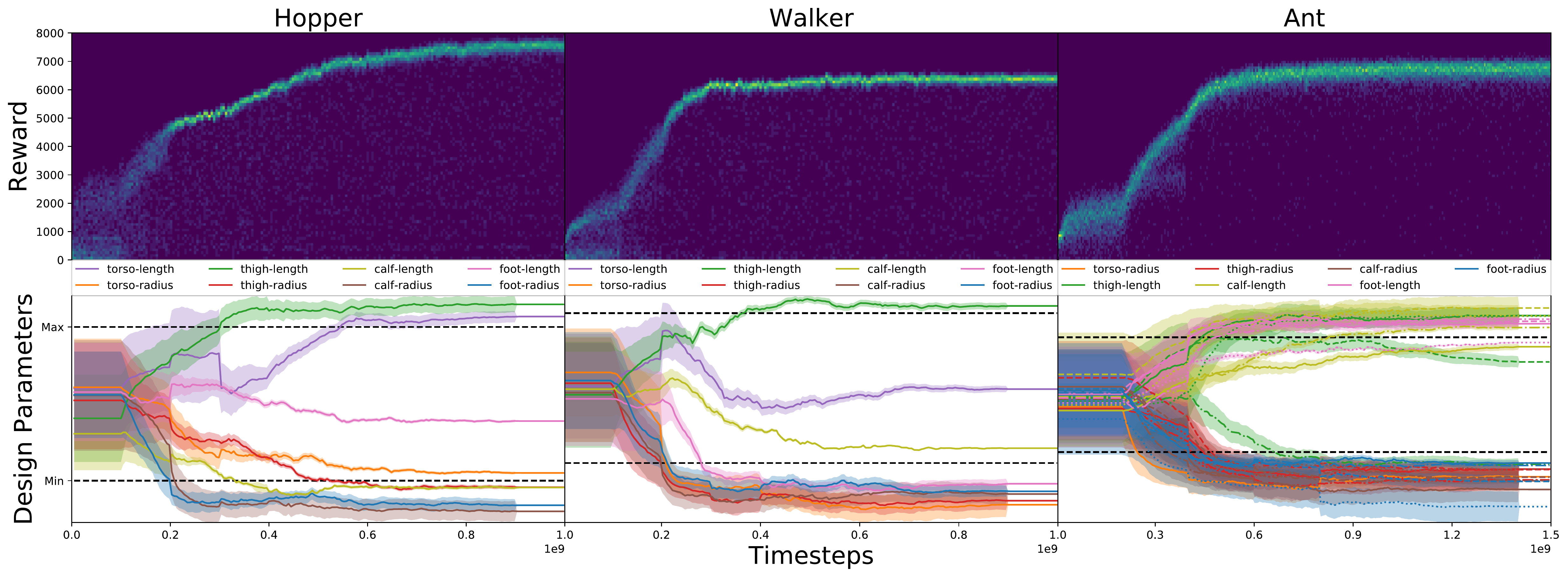}

    \caption{Histograms that show the reward evolution (top) for sampled designs of the Hopper, Walker, and Ant, along with the evolution of the mean and variance of the design distribution (bottom) during training. Every $5$\,M timesteps, we sample $100$ robots from our current design distribution and plot the distribution of rewards obtained by those robots over the course of an episode. In the plot showing design parameters for the Ant, colors represent the different parts of each leg and line styles represent individual legs.} \label{fig:histogram-params-flat}
\end{figure*}

\subsubsection*{Inclined Terrain} On the inclined terrain, our method again outperforms the baselines for the Hopper and Walker morphologies.  For the Hopper, there is again some variance in the reward achieved by our method among the different seeds, but the learned designs are fairly consistent. The learned Hopper designs again have long, thin bodies and a long foot. The long foot allows the robot to easily move up the incline by gaining height while pushing off the front of the foot. The Walker has more variance than in the level terrain case---while all of the learned designs have long thin bodies, we observe two distinct gaits. One is a standard walking gait and the other always keeps one foot in front of the other, pushing off with both feet simultaneously. Surprisingly, the latter gait receives higher reward. Without the need to move one leg past the other, the Walker can efficiently move up the incline, reducing the penalty for applying torques at each joint.

Standard training with the Ant on inclined terrain represents an interesting failure case for our method. While one of our seeds outperforms all but one baseline seed, for the remaining seven, the optimization remains stuck at a local optimum with very low reward---corresponding to the Ant just balancing itself on the incline without moving forward. We believe this is the case because the control network is unable to make a significant portion of the sampled designs move forward during the initial warm-up period (when the design is fixed), and chooses the ``safer'' option of having all samples stand still. The optimization is never able to move away from this local optimum once the design distribution starts updating, moving towards large designs that balance using little energy, making locomotion even harder.

While an interesting direction for future work is finding a more general solution for avoiding local optima, we find that a simple modification to the reward function suffices in this case. During training, we reduce the reward for staying upright by half, thus further incentivizing forward motion. We include these results in Fig.~\ref{fig:rewards} (also, the included result in Fig.~\ref{fig:evolution-gaits} corresponds to the best seed trained with this modified reward). With this modification, all eight seeds are able to escape the standing local minima, and all outperform the baselines---with performance measured in terms of the original reward. (However, note that a similar modification during training may also have helped the baselines.)

\subsection{Evolution of Design and Control During Training}

To better understand the training process and the evolution of our learned designs, we provide a histogram of episode rewards throughout training, along with the marginal distributions for each design parameter, for different runs of our experiments (Fig.~\ref{fig:histogram-params-flat}). The histogram is generated as follows: every $5$\,M timesteps during training, we sample $100$ designs from the design distribution and report the episode reward of each design under the current control policy. Figure~\ref{fig:histogram-params-flat} shows that our approach maintains high variance the in design parameters early in training, then slowly converges to a narrow distribution before being fixed at the mode of the distribution at the end of training. Not surprisingly, the evolution of the reward histograms closely mimic changes to the means and variances of the design distributions. There are clear shifts in the design distributions and the reward histograms when components are evaluated and pruned.

\section{Conclusion} \label{sec:conclusion}

We proposed a model-free algorithm that jointly optimizes over a robot's physical design and the corresponding control policy, without the need for any expert supervision. Given an arbitrary morphology, our method maintains a distribution over the physical design parameters together with a neural network control policy. This control policy has access to design parameters, allowing it to generalize policies to novel parts of design space. Using reinforcement learning, our method updates the control policy to maximize the expected reward over the design distribution while simultaneously shifting the distribution towards higher-performing designs. The method thereby converges to a design and policy that are locally optimal. We evaluated our approach on a variety legged robot morphologies for different locomotion tasks, demonstrating that our method results in novel robot designs and walking gaits that outperform baselines based on Bayesian optimization and randomly sampling designs.

Our findings suggest several avenues for future work. The most direct is extending the current approach to find optimized designs for a larger variety of tasks, such as locomotion in the presence of uneven terrain, obstacles, variations in friction, etc; and other domains such as manipulation. We are also interested in extending our framework to relax the assumption that the morphology is fixed. This would require a more complicated policy capable of handling different action spaces for each morphology within the design distribution.
Another avenue for future work is to fabricate the learned design and  transfer the control policy from simulation to the real robot. Recent work by \citet{tan18} has shown this to be possible in the context of legged locomotion.

\bibliographystyle{IEEEtranN}

{\small
\bibliography{references}
}

\end{document}